\documentclass[preprint,3p,twocolumn]{elsarticle}

\usepackage{amssymb}
\usepackage{amsthm}
\usepackage[ruled,vlined]{algorithm2e}
\usepackage{algorithmic}
\usepackage{graphicx}
\usepackage{epsfig}
\usepackage{amsmath}
\usepackage{lscape}
\usepackage{tabularx}
\usepackage{color, soul}
\usepackage{makecell}
\usepackage[table]{xcolor}
\usepackage{colortbl}
\usepackage{rotating}
\usepackage{mathtools}
\usepackage[colorlinks = true,
            linkcolor = blue,
            urlcolor  = blue,
            citecolor = blue,
            anchorcolor = blue]{hyperref}
\usepackage[left]{lineno}

\usepackage{comment}

\makeatletter
\def\ps@pprintTitle{%
  \let\@oddhead\@empty
  \let\@evenhead\@empty
  \let\@oddfoot\@empty
  \let\@evenfoot\@oddfoot
}
\makeatother

\begin{document}

\begin{frontmatter}

\title{RRT-GPMP2: A Motion Planner for Mobile Robots in Complex Maze Environments}

\author[1,2]{Jiawei~Meng\corref{cor1}}
\ead{jiawei.meng@ucl.ac.uk}
\ead{jiaweimeng1994@gmail.com}
\author[1]{Danail~Stoyanov}
\cortext[cor1]{Corresponding author: Jiawei Meng}
\address[1]{Department of Computer Science, University College London, London WC1E 6BT, UK}
\address[2]{Department of Mechanical Engineering, University College London, London WC1E 6BT, UK}

\begin{abstract}
 With the development of science and technology, mobile robots are playing a significant important role in the new round of world revolution. Further, mobile robots might assist or replace human beings in a great number of areas. To increase the degree of automation for mobile robots, advanced motion planners need to be integrated into them to cope with various environments. Complex maze environments are common in the potential application scenarios of different mobile robots. This article proposes a novel motion planner named the rapidly exploring random tree based Gaussian process motion planner 2, which aims to tackle the motion planning problem for mobile robots in complex maze environments. To be more specific, the proposed motion planner successfully combines the advantages of a trajectory optimisation motion planning algorithm named the Gaussian process motion planner 2 and a sampling-based motion planning algorithm named the rapidly exploring random tree. To validate the performance and practicability of the proposed motion planner, we have tested it in several simulations in the Matrix laboratory and applied it on a marine mobile robot in a virtual scenario in the Robotic operating system.
\end{abstract}


\end{frontmatter}

\section{Introduction}

 \begin{table*}[t!]
 \caption{A summary of abbreviations in this article.}
 \label{Table:abbr}
 \centering
 \small
 \begin{tabular}{|m{2.5cm}|m{7.5cm}|}
 \hline
 \textbf{Abbreviations} & \textbf{Definitions} \\ \hline
 \cellcolor{green!25}2-D & \cellcolor{yellow!25}2 dimensional\\
 \cellcolor{green!25}3-D & \cellcolor{yellow!25}3 dimensional\\
 \cellcolor{green!25}GP & \cellcolor{yellow!25}Gaussian process \\
 \cellcolor{green!25}GPMP & \cellcolor{yellow!25}Gaussian process motion planner \\
 \cellcolor{green!25}LTV-SDE & \cellcolor{yellow!25}Linear time-varying stochastic differential equation\\
 \cellcolor{green!25}MAP & \cellcolor{yellow!25}Maximum a posterior\\
 \cellcolor{green!25}MATLAB & \cellcolor{yellow!25}Matrix laboratory \\
 \cellcolor{green!25}ROS & \cellcolor{yellow!25}Robotics operating system \\ 
 \cellcolor{green!25}RRT & \cellcolor{yellow!25}Rapidly exploring random tree \\
 \cellcolor{green!25}RRT-GPMP2 & \cellcolor{yellow!25}Rapidly exploring random tree based Gaussian process motion planner 2\\
\cellcolor{green!25}SDC & \cellcolor{yellow!25}Self-driving car\\
 \cellcolor{green!25}UAV & \cellcolor{yellow!25}Unmanned aerial vehicle\\
 \cellcolor{green!25}USV & \cellcolor{yellow!25}Unmanned surface vehicle\\
 \cellcolor{green!25}UVV & \cellcolor{yellow!25}Unmanned underwater vehicle\\
 \cellcolor{green!25}WAM-V 20 & \cellcolor{yellow!25}Wave adaptive modular vessel 20\\
 \hline
 \end{tabular}
 \end{table*}

 Mobile robots have been widely used in many different fields during the past years and are still developing rapidly nowadays \cite{rubio2019review}. This trend will probably remain for another decade based on our predictions. Depending on the scope of applications, mobile robots can be classified as aerial robots, ground robots, marine robots and others \cite{dji, limocorot, wamv}. Sometimes mobile robots might be deployed in maze environments to perform autonomous missions. For example, UAVs might require to enter complex, maze-like environments to perform autonomous rescue missions after disasters. Moreover, SDCs and USVs might require to perform autonomous parking missions in complex, maze-like garages or ports. 

 \begin{figure}[b!]
 \centering
 \includegraphics[width=1\linewidth]{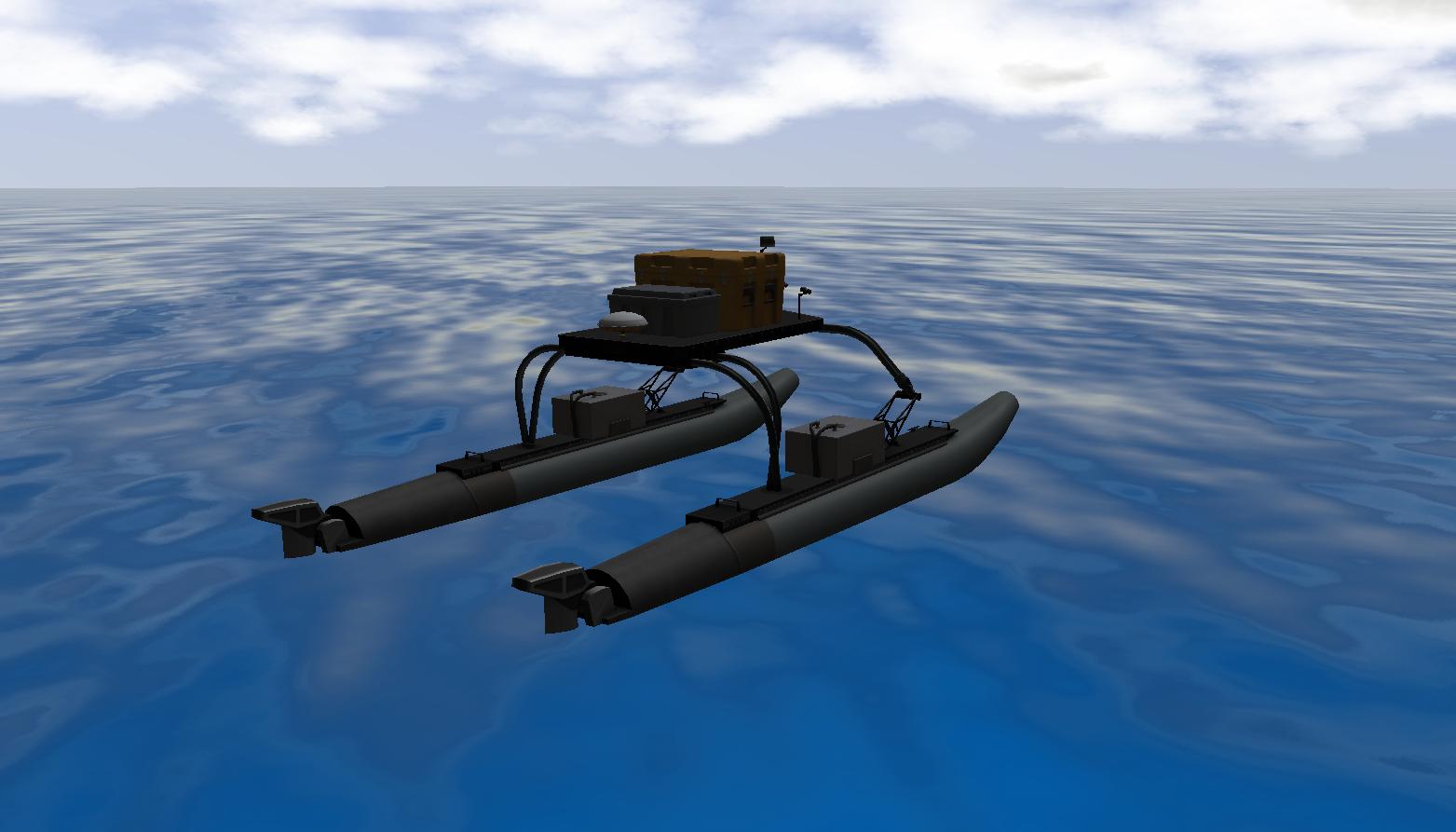}
 \centering
 \caption{Detailed demonstration of the selected USV platform in the Gazebo simulation environment in ROS \cite{bingham2019toward}. The USV platform is floating on the ocean surface and it will perform a path-following mission to validate our proposed motion planner.}
 \label{WAM_V} 
 \end{figure}
 
 Motion planning is an essential component to ensure mobile robots can successfully perform autonomous missions in maze-like environments. Based on the previous studies in \cite{meng2022anisotropic, meng2022fully}, we can divide the current mainstream motion planners into five categories and they are 1) Grid-based motion planners \cite{dijkstra2022note, hart1968formal, stentz1994optimal, sethian1996fast, lolla2012path}, 2) Sampling-based motion planners \cite{kavraki1996probabilistic, lavalle1998rapidly, meng2018uav, gammell2014informed, kuffner2000rrt}, 3) Potential field motion planners \cite{khatib1986real, petres2005underwater, garrido2008exploration}, 4) Intelligent motion planners \cite{colorni1991distributed, whitley1994genetic, mahmoudzadeh2016novel} and 5) Trajectory optimisation motion planners \cite{mukadam2016gaussian, barfoot2014batch, yan2017incremental, dong2016motion, mukadam2018continuous, dellaert2012factor}. The characteristics of these mainstream motion planners can be summarised as \cite{meng2022anisotropic, meng2022fully}: 
 \begin{itemize}
     \item Grid-based motion planners can always generate the shortest path. However, them require a long computational time to find the path, especially for high-dimensional or complex planning environments \cite{dijkstra2022note, hart1968formal, stentz1994optimal, sethian1996fast, lolla2012path}.
     \item Sampling-based motion planners can always generate a feasible path by exploring the planning environments due to their randomness. However, a relatively long computational time might be required and the smoothness of the path cannot be guaranteed. The post-planning smoothing process is required for this category of motion planners \cite{kavraki1996probabilistic, lavalle1998rapidly, meng2018uav, gammell2014informed, kuffner2000rrt}.
     \item Potential field motion planners and intelligent motion planners sometimes cannot find a feasible path due to the problem of local minima and usually require a relatively long computational time \cite{khatib1986real, petres2005underwater, garrido2008exploration, colorni1991distributed, whitley1994genetic, mahmoudzadeh2016novel}.
     \item Trajectory optimisation motion planners can find an optimised path in an efficient manner while smoothing the path in line with the planning process \cite{mukadam2016gaussian, barfoot2014batch, yan2017incremental, dong2016motion, mukadam2018continuous, dellaert2012factor}.
 \end{itemize}
 
 The performance of trajectory optimisation motion planners is more superior compared with the performances of other mainstream motion planners \cite{meng2022anisotropic, meng2022fully}. But the performance of trajectory optimisation motion planners might be limited in maze environments. There are some research about applying trajectory optimisation motion planners in simple maze environments \cite{petrovic2019stochastic, petrovic2020cross}. Nevertheless, there is still a lack of exploration of utilising them in complex maze environments. 

 In general, trajectory optimisation motion planners have a more powerful performance compared with other mainstream motion planners due to its superior computational speed and high-quality of the generated paths. Consequently, the proposed motion planner, RRT-GPMP2, uses one of the trajectory optimisation motion planners, GPMP2, as a fundamental means to generate a path in a complex maze environment. Additionally, one of the sampling-based motion planners, RRT, is used as an auxiliary means if the performance of GPMP2 is restricted in a part of the complex maze environment. In simple terms, the proposed motion planner, RRT-GPMP2, combines the advantages of trajectory optimisation motion planners and sampling-based motion planners.

 We have tested the proposed motion planner in two self-constructed, complex maze simulation environments in MATLAB to validate its performance. To validate the practicability of the proposed motion planner, we have also tested it on a USV platform in a high-fidelity, maritime simulation environment in ROS. Details of the selected USV platform in the simulation environment is provided in Fig. \ref{WAM_V}.

The contributions of this article are summarised as follows:
\begin{itemize}
    \item The RRT-GPMP2 motion planner is proposed to solve the motion planning problem in complex maze environments. Specifically, it effectively combines GPMP2 global planning and RRT local re-planning.
    \item The proposed RRT-GPMP2 motion planner is thoroughly tested in a series of simulations to validate its performance and practicability.
\end{itemize}

The rest of this article is organised as follows: Section \ref{Proposed motion planner} presents the proposed RRT-GPMP2 motion planner in detail. Section \ref{Simulations and discussions} demonstrates the simulation results of the proposed RRT-GPMP2 motion planner in MATLAB. Section \ref{Demonstration in ROS} demonstrates the practicability of the proposed RRT-GPMP2 motion planner in ROS. Section \ref{Conclusion} concludes this article as well as proposes some future research directions.

\section{Proposed motion planner}
\label{Proposed motion planner}
This section presents the proposed RRT-GPMP2 motion planner, in detail, which can be divided into two main parts:
\begin{itemize}
    \item The GPMP2 global planning.
    \item The RRT local re-planning.
\end{itemize}

To be more specific, we present the principle of GPMP2 in Section \ref{GPMP2}, the principle of RRT in Section \ref{RRT} and the combination of the GPMP2 global planning and the RRT local re-planning in Section \ref{Combination of the GPMP2 global planning and the RRT local re-planning}.

\subsection{GPMP2}
\label{GPMP2}
This subsection details the GPMP2 algorithm, which is used as the global planning part of the proposed RRT-GPMP2 motion planner. 

The GPMP2 algorithm was proposed in \cite{dong2016motion, mukadam2018continuous} and used in previous studies \cite{meng2022anisotropic, meng2022fully}. These studies are briefly summarised in this subsection to explain the principle of GPMP2.

Generally, the GP-based motion planning algorithms, such as GPMP2, view motion planning as a probabilistic inference process, which infers an optimised posterior based on the \textit{GP prior} and \textit{likelihood} distributions \cite{meng2022anisotropic, meng2022fully, dong2016motion, mukadam2018continuous}. The probabilistic inference process to obtain an optimised \textit{posterior} is called the MAP estimation and can be formulated as: $\theta^* = \mathop{\arg\max}_{\theta} \; p(\theta)l(\theta; e)$, where $\theta^*$ is the optimised \textit{posterior}, $p(\theta)$ is the \textit{GP prior} distribution and $l(\theta; e)$ is the \textit{likelihood} distribution. $p(\theta|e)$ can also be used to describe the optimised \textit{posterior}. The Levenberg-Marquardt method is used to solve the least squares problem during the optimisation process of GPMP2.

The \textit{GP prior} is a vector-valued GP that contains several continues-time trajectories samples: $\theta(t)\sim\mathcal{GP}(\mu(t),K(t,t'))$, where $\mu(t)$ is a vector-valued mean function and $K(t,t')$ is a matrix-valued covariance function. The \textit{GP prior} distribution can be expressed as: $p(\theta) \propto exp\{ -\frac{1}{2}||\theta - u||^{2}_{K}\}$. In GPMP2, a constant-velocity system dynamics model is used to generate the \textit{GP prior}.
 
The \textit{likelihood} describes the probability of having a collision with any obstacle at each point in the environment and it can be expressed as: $l(\theta; e) = exp\{ -\frac{1}{2}||h(\theta)||^{2}_{\Sigma_{obs}}\}$, where $\Sigma_{\textit{obs}}$ is a hyperparameter matrix and $h(\theta)$ is a vector-valued obstacle cost function: $h(\theta_{i}) = [c(d(\theta_{i}))]$, where $c$ is a hinge loss function and $\textit{d}$ is the signed distance of a point.

Factor graphs are used as an efficient optimisation tool to deal with the probabilistic inference in MAP estimation in the GPMP2 algorithm. The major difference between GPMP, the original version of GP-based motion planning, and GPMP2, an improved version of GPMP, is that GPMP2 employs a factor graph to reduce the time cost during the optimisation process. A factor graph can be formulated as: $G = \{\Theta, \mathcal{F}, \mathcal{E}\}$, where $\Theta=\{ \theta_{0},...,\theta_{N} \}$ are \textit{variable nodes}, $\mathcal{F}=\{f_{0},...,f_{M}\}$ are \textit{factor nodes} and $\mathcal{E}$ are \textit{edges} that connect the \textit{variable nodes} and \textit{factor nodes}. The posterior factorisation can be expressed as: $p(\theta|e) \propto \prod^{M}_{m=1}f_{m}(\Theta_{m})$, where $f_{m}$ are a series of factors on variable subset $\Theta_{m}$ \cite{meng2022anisotropic, meng2022fully, dong2016motion, mukadam2018continuous}.

\subsection{RRT}
\label{RRT}
This subsection details the RRT algorithm, which is used as the local planning part of the proposed RRT-GPMP2 motion planner. 

The RRT algorithm was proposed in \cite{lavalle1998rapidly} and further explained in \cite{lavalle2001randomized, lavalle2006planning, karaman2011sampling, gao2022bim}. These studies are briefly summarised in this subsection to explain the principle of RRT. 

 \begin{algorithm}[t!]
 \SetAlgoLined
 \textbf{Input:} Start point $x_{start}$ and goal point $x_{goal}$\\
 
 \For{$i=1,2,\ldots,N$}{
            $\textit{x}_{rand} \leftarrow$ \textbf{Sample($X_{free}, N$)};\\
            $\textit{x}_{nearest} \leftarrow$ \textbf{Nearest($\textit{V}, x_{rand}$)};\\
            $\textit{x}_{new} \leftarrow$ \textbf{Steer($\textit{x}_{nearest}, x_{rand}$)};\\
            \If{\rm\textbf{{CollisionFree}($\textit{x}_{nearest}, \textit{x}_{new}$)}}
            {
                $\textit{V} \leftarrow$ $\textit{V}$ $\cup$ {$x_{new}$};\\
                $\textit{E} \leftarrow$ $\textit{E}$ $\cup$ {($x_{nearest}$, $x_{new}$)};\\
		    }}
      
 \textbf{Output:} Tree structure $T = (V, E)$\\
 \caption{RRT Algorithm \cite{lavalle1998rapidly, lavalle2001randomized, lavalle2006planning, karaman2011sampling, gao2022bim}}
 \label{alg:rrt}
 \end{algorithm}
 
Generally, the RRT algorithm generates a tree structure from a start point to a goal point in a space while avoiding any obstacles \cite{lavalle1998rapidly, lavalle2001randomized, lavalle2006planning, karaman2011sampling, gao2022bim}. The generated tree structure can be viewed as a collision-free path $T=(V, E)$, where $V$ represents the \textit{vertices} on the tree structure and $E$ represents the \textit{edges} on the tree structure. The entire metric space is represented as $X$, the region with static obstacles is represented as $X_{obs} \subset X$ and the region without static obstacles is represented as $X_{free} \subset X$. The start point of the motion planning problem is represented as $x_{start} \in X_{free}$ and the goal point is represented as $x_{goal} \in X_{free}$. The general process of the RRT algorithm to construct a tree structure is demonstrated in Algorithm \ref{alg:rrt} and it can be summarised as follows:

 \begin{itemize}
     \item First, if the tree structure $T = (V, E)$ has not reached the goal point $x_{goal}$ yet, \textbf{Sample($X_{free}, N$)} generates a random point $x_{rand}$ inside the region without static obstacles $X_{free}$.
     \item Second, \textbf{Nearest($\textit{V}, x_{rand}$)} performs a comparison between the randomly sampled point $x_{rand}$ and the rest states in the set of nodes $V$ to find the nearest point $x_{nearest}$ to $x_{rand}$.
     \item Third, \textbf{Steer($x_{nearest}, x_{rand}$)} generates a new point $x_{new}$, which is closer to $x_{nearest}$ by connecting $x_{rand}$ and $x_{nearest}$ with a steering function.
     \item Fourth, \textbf{CollisionFree($x_{nearest}, x_{new}$)} checks if there is any collision between the straight path from $x_{new}$ to $x_{nearest}$ and the region with static obstacles $X_{obs}$;
     \item Fifth, the new point $x_{new}$ is added to the set of nodes $V$ and the new edges that connects $x_{rand}$ and $x_{new}$ is added to the set of edges $E$;
     \item Last, the mentioned processes repeat for $N$ times until the tree structure $T = (V, E)$ reaches the goal point $x_{goal}$.
 \end{itemize}

 \begin{figure}[t!]
 \centering
 \includegraphics[width=1\linewidth]{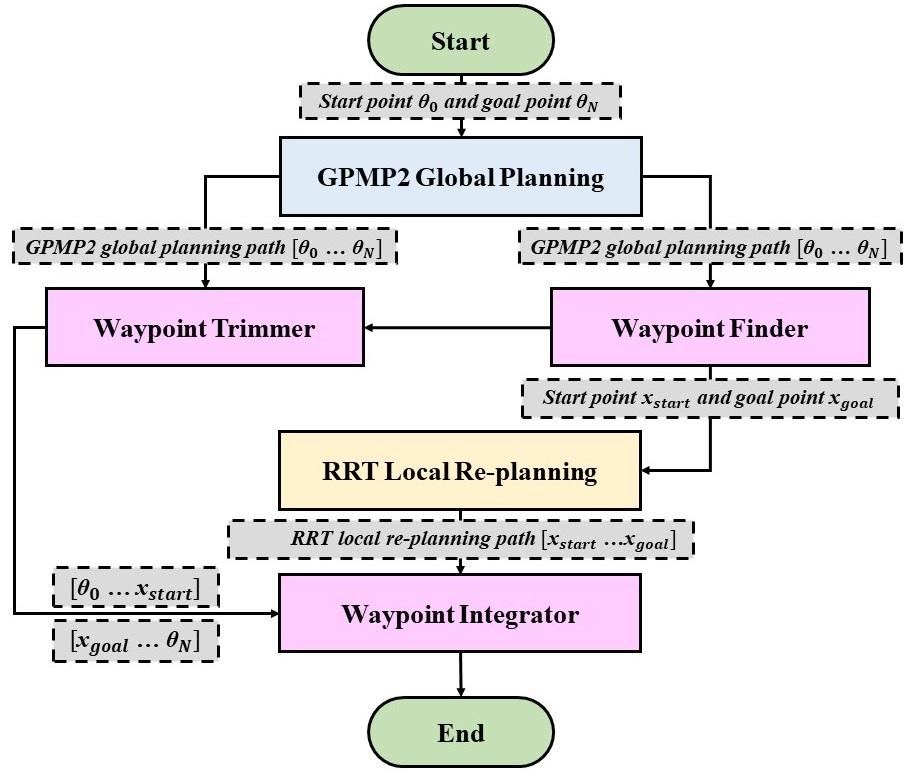}
 \centering
 \caption{Flow diagram of the proposed RRT-GPMP2 motion planner.}
 \label{RRT-GPMP2-flowchart} 
 \end{figure} 

\subsection{Combination of the GPMP2 global planning and the RRT local re-planning}
\label{Combination of the GPMP2 global planning and the RRT local re-planning}
This subsection details the proposed RRT-GPMP2 motion planner, which successfully combines the GPMP2 global planning and the RRT local re-planning in an effective manner.

The GPMP2 algorithm can generate a path between the start point and goal point in an extremely short time due to \cite{meng2022anisotropic, meng2022fully, dong2016motion, mukadam2018continuous}:
\begin{itemize}
    \item The guidance of the mean and covariance of the GP prior samples.
    \item The efficiency of the incremental optimisation tool (factor graph).
\end{itemize}
Nevertheless, the path generated by the GPMP2 algorithm might not satisfy the collision-free requirement in a complex maze environment. 

Compared with the GPMP2 algorithm, the RRT algorithm can always generate a feasible path in a complex maze environment. But the RRT algorithm usually requires a relatively long computational time and the smoothness of the generated path is low. 

Thus, we propose the RRT-GPMP2 motion planner, which uses the GPMP2 algorithm described in Section \ref{GPMP2} to generate a global path and allows collisions to occur on it and then uses the RRT algorithm described in Section \ref{RRT} to generate a local collision-free path to substitute the collision part of the global path. 

To provide a more intuitive understanding of the proposed RRT-GPMP2 motion planner, its flow diagram is provided in Fig. \ref{RRT-GPMP2-flowchart} and its detailed process can be summarised as follows:

\begin{enumerate}
    \item The \textbf{GPMP2 Global Planning} generates a path [$\theta_{0}$...$\theta_{N}$] based on start point $\theta_{0}$ and goal point $\theta_{N}$.
    \item The \textbf{Waypoint Finder} defines the start point $x_{start}$ and goal point $x_{goal}$ of the RRT local re-planning by traversing the waypoints of the generated global path [$\theta_{0}$...$\theta_{N}$] and finds the collision part of the global path [$\theta_{a}$...$\theta_{b}$]. More specifically, $\theta_{a}$ is defined as $x_{start}$ and $\theta_{b}$ is defined as $x_{goal}$.
    \item The \textbf{RRT Local Re-planning} generates a local path [$x_{start}$...$x_{goal}$] based on the start point $x_{start}$ and goal point $x_{goal}$. 
    \item The \textbf{Waypoint Trimmer} removes the collision part of the global path [$\theta_{a}$...$\theta_{b}$], while remains the collision-free parts [$\theta_{0}$...$\theta_{a}$] and [$\theta_{b}$...$\theta_{N}$] on the global path.
    \item The \textbf{Waypoint Integrator} integrates the collision-free parts [$\theta_{0}$...$\theta_{a}$] and [$\theta_{b}$...$\theta_{N}$] on the global path and the local path [$x_{start}$...$x_{goal}$].
\end{enumerate}

A crucial aspect of the proposed RRT-GPMP2 motion planner is to determine the start point and goal point of the RRT local re-planning. To adequately address this crucial aspect, the \textbf{Waypoint Finder} traverses the global path to find the last waypoint before collision as the start point of local re-planning and the first waypoint after collision as the goal point of local re-planning.

\section{Simulations and discussions}
\label{Simulations and discussions}
In this section, the proposed RRT-GPMP2 motion planner is tested in two complex maze environments to validate its performance.

\subsection{Simulation details}
We use the proposed RRT-GPMP2 motion planner and RRT global planning to generate paths and record results for each motion planning problem. The proposed RRT-GPMP2 motion planner is designed on the basis of the open-source repositories in \cite{Mukadam-IJRR-18, Dong-RSS-16, dong2018sparse, rrt-code-2014} and the RRT (global planning and local re-planning) implemented is on the basis of the open-source repository in \cite{rrt-code-2014}. The parameters of the GPMP2 global planning are chosen based on our past experience. For both the motion planning problems in the different maze environments, the step length of RRT (global planning and local re-planning) is defined as 10 [m].

Simulations are run on a computer with 16 2.3-GHz Intel Core i7-11800H processors and a 8 GB RAM.

\subsection{First motion planning problem}
The start and goal positions are [400,  400] and [400,  100], respectively. Furthermore, the start and goal positions when the proposed RRT-GPMP2 motion planner starts to re-planning the path are [374, 286] and [367, 222], respectively. 

The detailed GPMP2 global planning process of the proposed RRT-GPMP2 motion planner in the first motion planning problem is demonstrated in Fig. \ref{RRT-GPMP2} (a), while the detailed RRT local re-planning process of the proposed RRT-GPMP2 motion planner in the first motion planning problem is demonstrated in Fig. \ref{RRT-GPMP2} (b). Moreover, the detailed RRT global planning process in the first motion planning problem is demonstrated in Fig. \ref{RRT-GPMP2} (c). 

Table \ref{Table:rrt-gpmp2-1} summarises the simulation results of the proposed RRT-GPMP2 motion planner in the first motion planning problem, while Table \ref{Table:rrt-1} summarises the simulation results of the RRT global planning in the first motion planning problem.

As demonstrated in Table \ref{Table:rrt-gpmp2-1}, the GPMP2 global planning process of the proposed RRT-GPMP2 motion planner can generate a global path in an extremely short time. Specifically, this process costs 27.1 [ms] and is only 0.52 [\%] of the total time cost. On the other hand, the total time cost of the RRT local re-planning process of the proposed RRT-GPMP2 motion planner costs 5151.6 [ms] and is the majority of the total time cost. This is because the RRT explores the maze environment without bias on the searching direction. It is beneficial for the RRT to find a feasible path. However, this diffuse search is time-consuming. The total time cost of the RRT global planning is 8592.6 [ms] as demonstrated in Table \ref{Table:rrt-1} and it is a significantly higher than the total time cost of the proposed RRT-GPMP2 motion planner. Moreover, the path lengths of the proposed RRT-GPMP2 motion planner and RRT global planning are similar according to Table \ref{Table:rrt-gpmp2-1} and Table \ref{Table:rrt-1}. Based on Fig. \ref{Comparison_total_time_cost} and Fig. \ref{Comparison_path_length}, we know that the proposed RRT-GPMP2 motion planner largely decreases the total time cost and maintains a similar path length of the generated path compared with the RRT global planning. The total time cost of the proposed RRT-GPMP2 motion planner is 60.27 [\%] of the total time cost of the RRT global planning. The path length of the proposed RRT-GPMP2 motion planner is 104.98 [\%] of the path length of the RRT global planning.

 \begin{table}[t!]
 \caption{Simulation results of the RRT-GPMP2 motion planner in the first motion planning problem.}
 \label{Table:rrt-gpmp2-1}
 \centering
 \footnotesize
 \begin{tabular}{|c|c|c|}
 \hline
 \textbf{Number} & \textbf{Measurement} & \textbf{Value} \\ \hline
  1 & \cellcolor{green!25}GPMP2 time cost & \cellcolor{yellow!25}27.1 [ms] \\ \hline
  2 & \cellcolor{green!25}RRT time cost & \cellcolor{yellow!25}5151.6 [ms] \\ \hline
  3 & \cellcolor{green!25}Total time cost & \cellcolor{yellow!25}\textbf{5178.7 [ms]} \\ \hline
  4 & \cellcolor{green!25}Path length & \cellcolor{yellow!25}977.5 [m] \\ \hline
\multicolumn{3}{p{200pt}}{\textbf{{Notes:}} {The total time cost in this table only includes the global planning time cost of GPMP2 and the local re-planning time cost of RRT.}}
 \end{tabular}
 \end{table}

 \begin{table}[t!]
 \caption{Simulation results of the RRT global planning in the first motion planning problem.}
 \label{Table:rrt-1}
 \centering
 \footnotesize
 \begin{tabular}{|c|c|c|}
 \hline
 \textbf{Number} & \textbf{Measurement} & \textbf{Value} \\ \hline
  1 & \cellcolor{green!25}Total time cost & \cellcolor{yellow!25}8592.6 [ms] \\ \hline
  2 & \cellcolor{green!25}Path length & \cellcolor{yellow!25}\textbf{931.1 [m]} \\ \hline
 \end{tabular}
 \end{table}

\subsection{Second motion planning problem}
In this motion planning problem, the start position are [250,  400] and the goal position are [400,  100]. The start and goal positions when the proposed RRT-GPMP2 motion planner starts to use the RRT algorithm to re-planning the path are [401, 176] and [400, 120], respectively. 

\begin{figure*}[t!]
 \centering
 \includegraphics[width=0.95\linewidth]{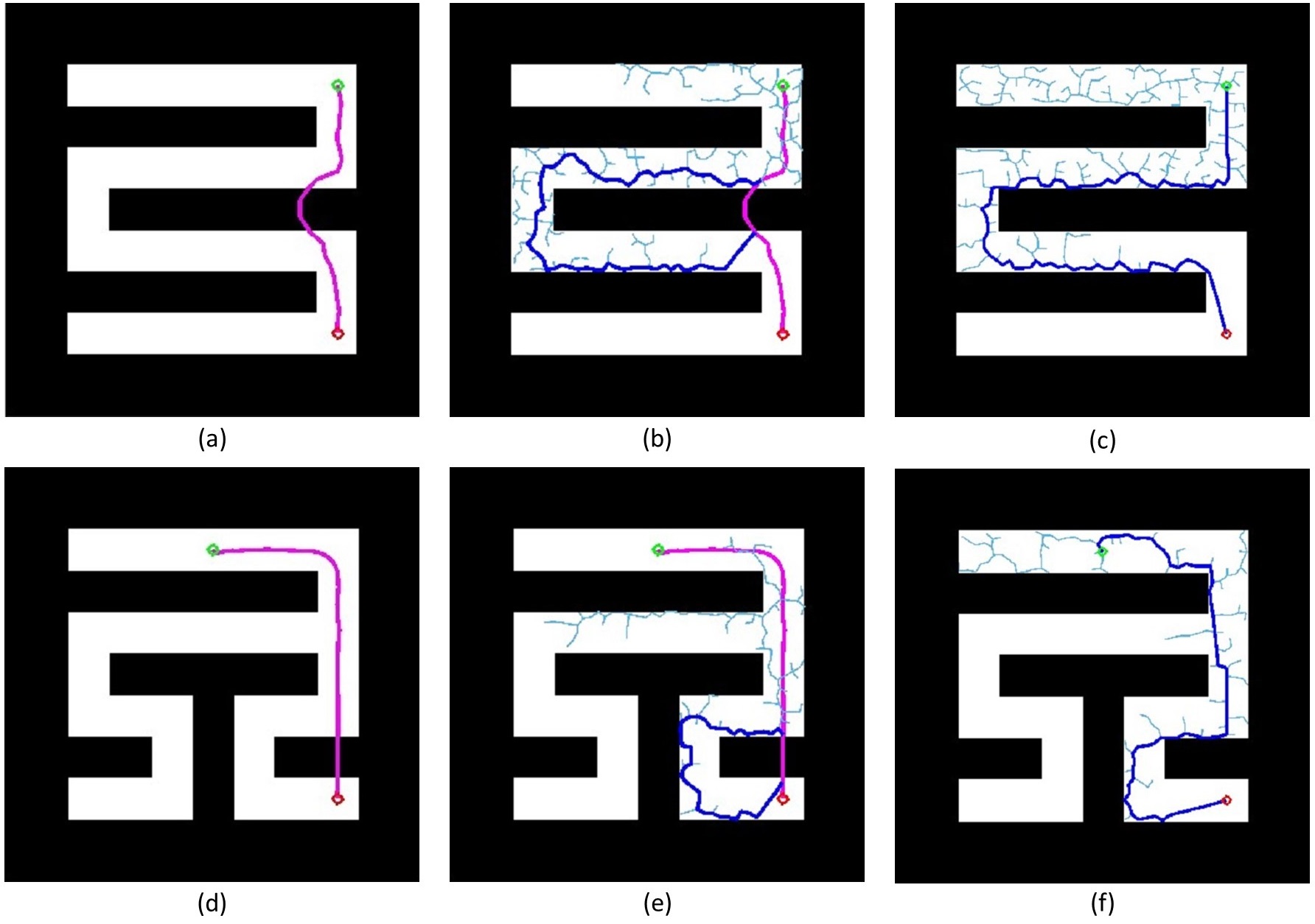}
 \centering
 \caption{Demonstration of the proposed RRT-GPMP2 motion planner and RRT global planning in the first and second motion planning problems \cite{Mukadam-IJRR-18, Dong-RSS-16, dong2018sparse, rrt-code-2014}. Sub-figures (a) and (b) demonstrate the GPMP2 global planning and the RRT local re-planning in respective in the first motion planning problem. Sub-figure (c) demonstrate the RRT global planning in the first motion planning problem. Sub-figures (d) and (f) demonstrate the GPMP2 global planning and the RRT local re-planning in respective in the second motion planning problem. Sub-figure (f) demonstrate the RRT global planning in the second motion planning problem. The start position is represented in green, the goal position is represented in red, the GPMP2 generated path is represented in purple, the RRT generated path is represented in blue and the RRT tree branches are represented in cyan.}
 \label{RRT-GPMP2} 
 \end{figure*}

The detailed GPMP2 global planning process of the proposed RRT-GPMP2 motion planner in the second motion planning problem is demonstrated in Fig. \ref{RRT-GPMP2} (d), while the detailed RRT local re-planning process of the proposed RRT-GPMP2 motion planner in the second motion planning problem is demonstrated in Fig. \ref{RRT-GPMP2} (e). Additionally, the detailed RRT global planning process in the second motion planning problem is demonstrated in Fig. \ref{RRT-GPMP2} (f). 

Table \ref{Table:rrt-gpmp2-2} provides a summary of the simulation results of the proposed RRT-GPMP2 motion planner in the second motion planning problem, while Table \ref{Table:rrt-2} provides a summary of the simulation results of the RRT global planning in the second motion planning problem.

 \begin{table}[t!]
 \caption{Simulation results of the RRT-GPMP2 motion planner in the second motion planning problem.}
 \label{Table:rrt-gpmp2-2}
 \centering
 \footnotesize
 \begin{tabular}{|c|c|c|}
 \hline
 \textbf{Number} & \textbf{Measurement} & \textbf{Value} \\ \hline
  1 & \cellcolor{green!25}GPMP2 time cost & \cellcolor{yellow!25}40.2 [ms] \\ \hline
  2 & \cellcolor{green!25}RRT time cost & \cellcolor{yellow!25}2574.8 [ms] \\ \hline
  3 & \cellcolor{green!25}Total time cost & \cellcolor{yellow!25}\textbf{2615.0 [ms]} \\ \hline
  4 & \cellcolor{green!25}Path length & \cellcolor{yellow!25}779.9 [m] \\ \hline
\multicolumn{3}{p{200pt}}{\textbf{{Notes:}} {The total time cost in this table only includes the global planning time cost of GPMP2 and the local planning time cost of RRT.}}
 \end{tabular}
 \end{table}

 \begin{table}[t!]
 \caption{Simulation results of the RRT global planning in the second motion planning problem.}
 \label{Table:rrt-2}
 \centering
 \footnotesize
 \begin{tabular}{|c|c|c|}
 \hline
 \textbf{Number} & \textbf{Measurement} & \textbf{Value} \\ \hline
  1 & \cellcolor{green!25}Total time cost & \cellcolor{yellow!25}4068.3 [ms] \\ \hline
  2 & \cellcolor{green!25}Path length & \cellcolor{yellow!25}\textbf{709.6 [m]} \\ \hline
 \end{tabular}
 \end{table}

 \begin{figure}[t!]
 \centering
 \includegraphics[width=0.95\linewidth]{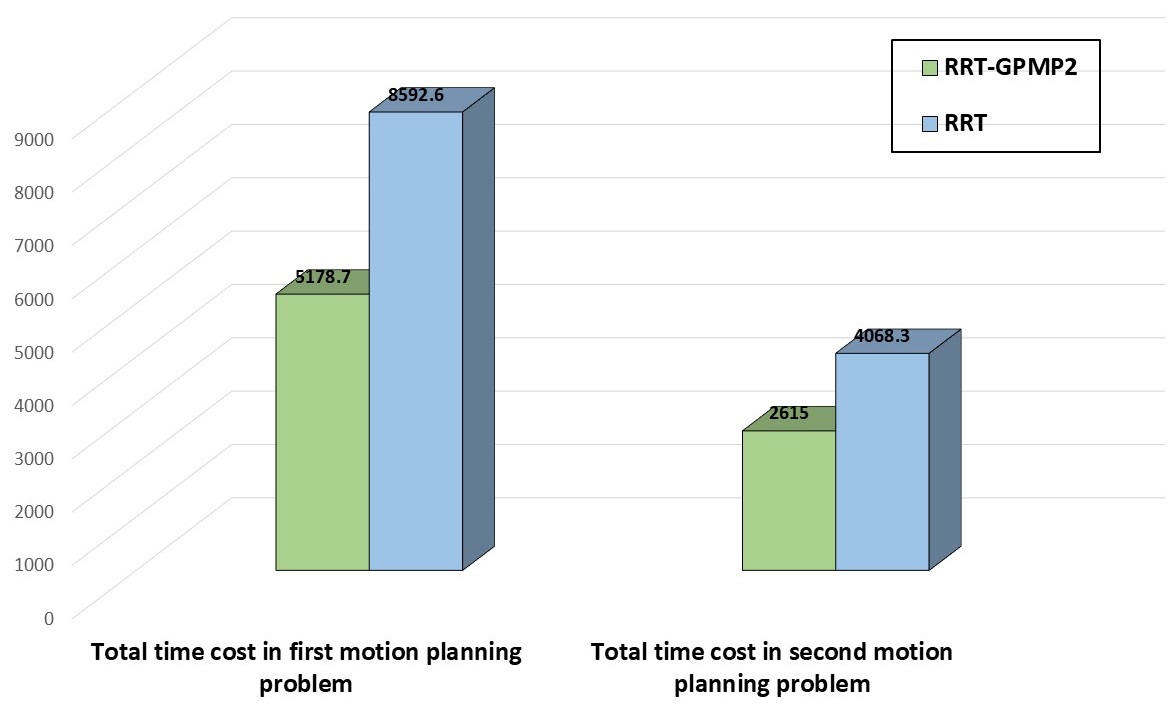}
 \centering
 \caption{Comparisons of the total time cost between the proposed RRT-GPMP2 motion planner and RRT global planning in both motion planning problems. Data in the figure is measured in milliseconds.}
 \label{Comparison_total_time_cost} 
 \end{figure}

 \begin{figure}[t!]
 \centering
 \includegraphics[width=0.95\linewidth]{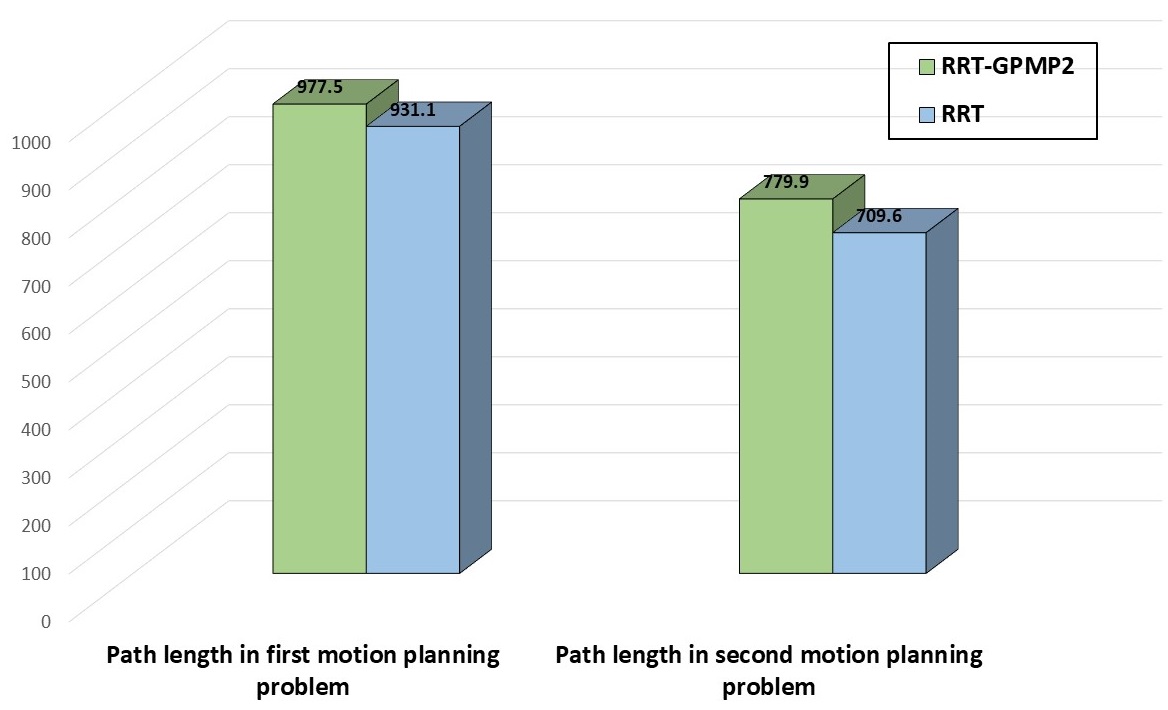}
 \centering
 \caption{Comparisons of the path length between the proposed RRT-GPMP2 motion planner and RRT global planning in both motion planning problems. Data in the figure is measured in meters.}
 \label{Comparison_path_length} 
 \end{figure}

 \begin{figure*}[t!]
 \centering
 \includegraphics[width=1.0\linewidth]{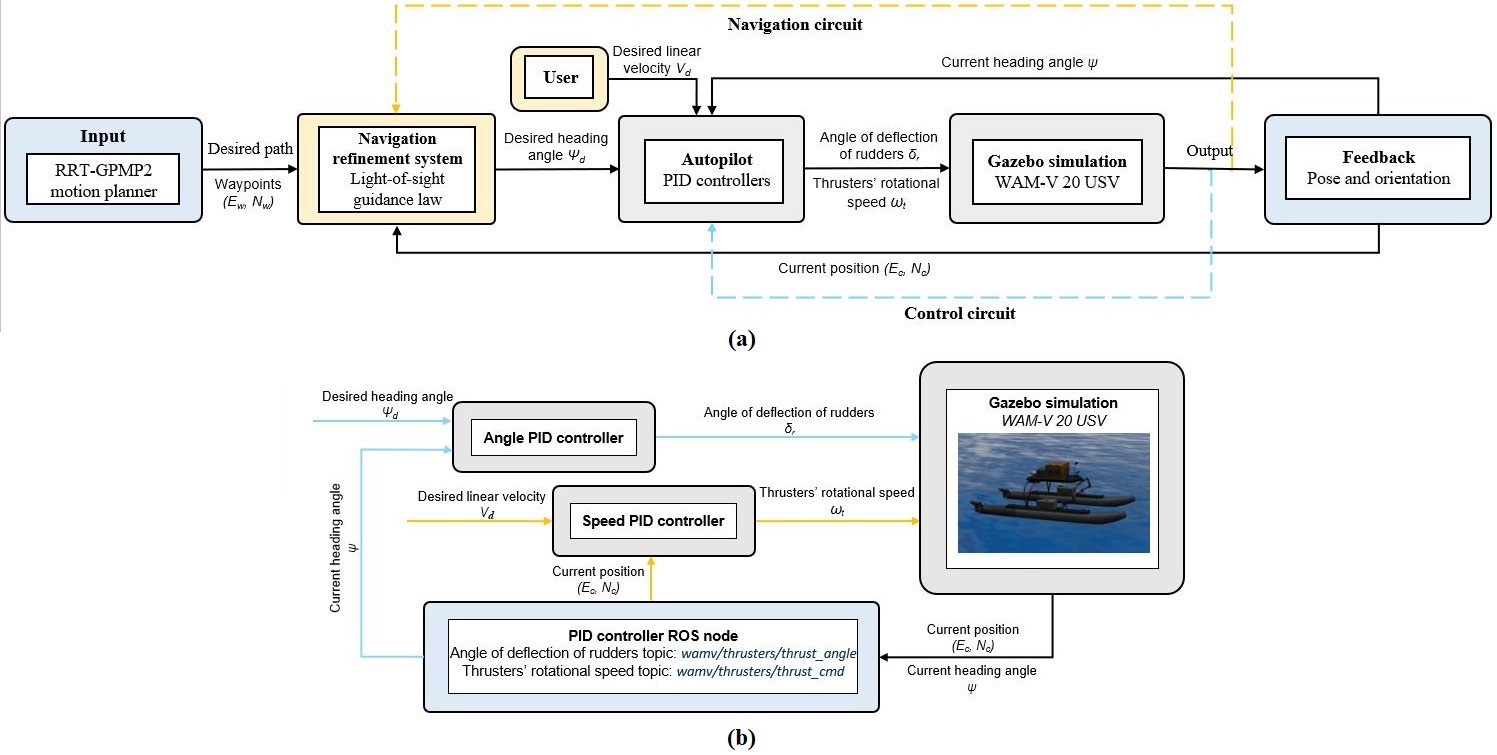}
 \centering
 \caption{(a) Overall structure of the entire system that includes the proposed RRT-GPMP2 motion planner and the fully-autonomous USV navigation framework proposed in \cite{meng2022fully}. (b) Detailed structure of the two controllers in the autopilot in sub-figure (a) \cite{bingham2019toward, meng2022fully}.}
 \label{fully_autonomous_system_RRT_GPMP2} 
 \end{figure*}

 \begin{figure*}[t!]
 \centering
 \includegraphics[width=0.8\linewidth]{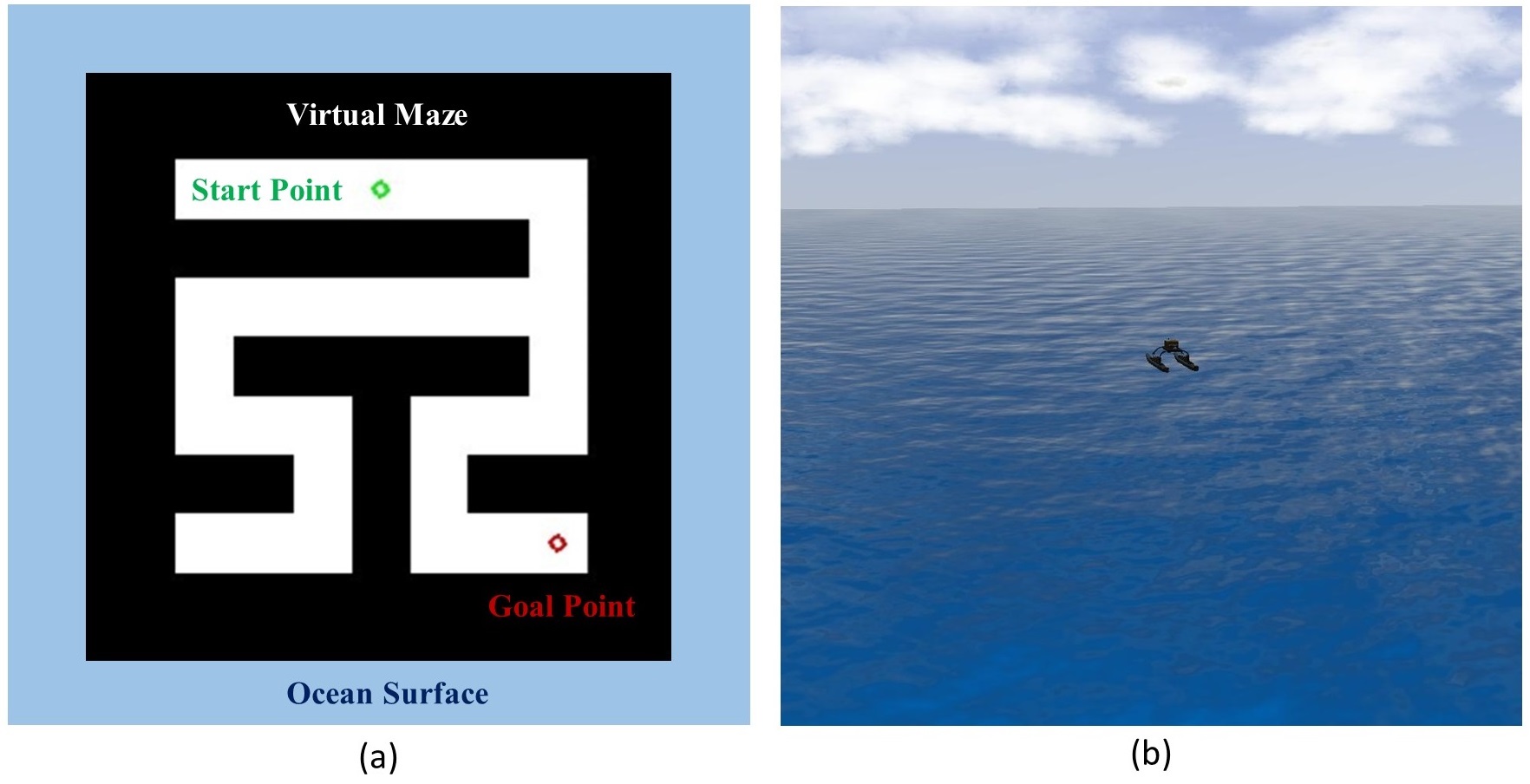}
 \centering
 \caption{(a) Path-following mission layout. (b) ROS simulation environment overview of the path-following mission \cite{bingham2019toward}.}
 \label{ros_layout_and_demonstration} 
 \end{figure*}

 \begin{figure*}[t!]
 \centering
 \includegraphics[width=1.0\linewidth]{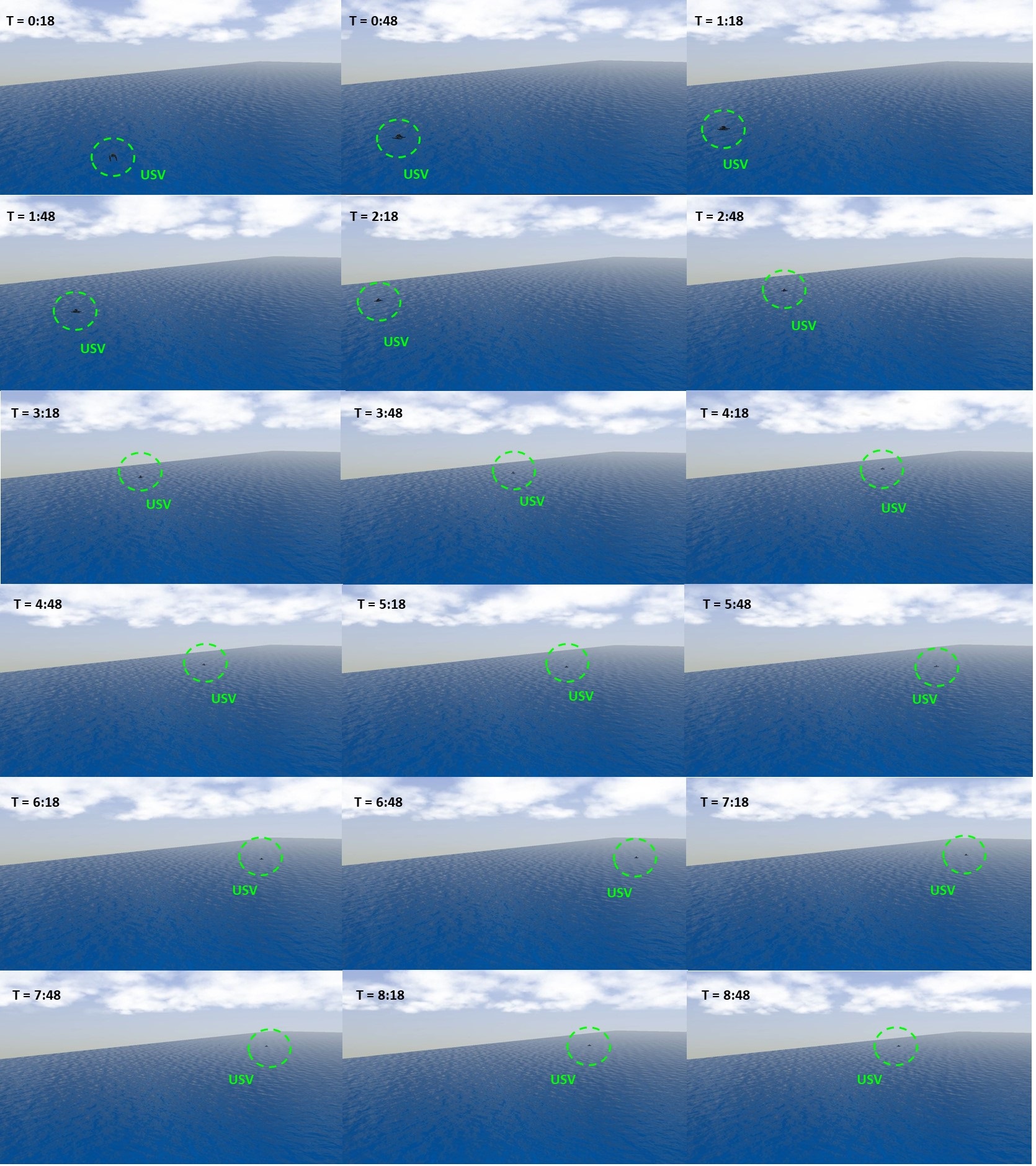}
 \centering
 \caption{The storyboards of the path-following mission using the proposed RRT-GPMP2 algorithm. Specifically, the storyboards record the state of the USV from 0 [min] 18 [sec] to 8 [min] 48 [sec]. The USV moves towards the start point from a nearby point between 0 [min] 0 [sec] and 0 [min] 18 [sec]. The USV roughly reaches the goal point at 8 [min] 48 [sec]. As mentioned earlier, this simulation was conducted in a ROS simulation environment based on the previous work in \cite{bingham2019toward, meng2022fully}.}
 \label{ros_path_following} 
 \end{figure*}

The GPMP2 global planning process of the proposed RRT-GPMP2 motion planner in this motion planning problem is 40.2 [ms] as demonstrated in Table \ref{Table:rrt-gpmp2-2} and is 1.54 [\%] of the total time cost of the proposed RRT-GPMP2 motion planner. It is similar to the situation in the first motion planning problem, the total time cost of the RRT local re-planning process is still the majority of the total time cost of the proposed RRT-GPMP2 motion planner. The total time cost of the RRT global planning is 4068.3 [ms] as demonstrated in Table \ref{Table:rrt-2} and it is still much larger than the total time cost of the proposed RRT-GPMP2 motion planner. Similar to the situation in the first motion planning problem again, the proposed RRT-GPMP2 and RRT have similar path lengths based on Table \ref{Table:rrt-gpmp2-2} and Table \ref{Table:rrt-2}. Based on Fig. \ref{Comparison_total_time_cost} and Fig. \ref{Comparison_path_length}, we can see that the proposed RRT-GPMP2 motion planner significantly reduces the total time cost and maintains a similar path length compared with the RRT global planning. More specifically, the total time cost of the proposed RRT-GPMP2 motion planner is 64.28 [\%] of the total time cost of the RRT global planning. The path length of the proposed RRT-GPMP2 motion planner is 109.91 [\%] of the path length of the RRT global planning. As shown in Fig. \ref{RRT-GPMP2}, the majority of the path generated by the proposed RRT-GPMP2 motion planner is smooth, but the path generated by the RRT global planning is zigzag from the start point to the goal point. The proposed RRT-GPMP2 motion planner just have a portion of the path that is zigzag. Because the GPMP2 global planning can always generate a smooth path. The proposed RRT-GPMP2 can guarantee part of the generated path is smooth. However, the RRT global planning cannot guarantee smoothness on any part of the path due to its randomness.

\section{Demonstration in ROS}
\label{Demonstration in ROS}
The marine mobile robots, such as USVs and UUVs, have gained increasing attention in the past decade. Consequently, this section tested the proposed motion planner in a fully-autonomous USV navigation framework in a high-fidelity ROS maritime environment to showcase its practicability \cite{meng2022fully}. Furthermore, this framework was proposed in a previous study in \cite{meng2022fully} and it was designed based on an open-source repository provided in \cite{bingham2019toward}. The overall structure of the entire system including the proposed RRT-GPMP2 motion planner and the framework is demonstrated in Fig. \ref{fully_autonomous_system_RRT_GPMP2} (a), while the detailed structure of the two controllers in the framework is demonstrated in Fig. \ref{fully_autonomous_system_RRT_GPMP2} (b).

 \begin{figure}[t!]
 \centering
 \includegraphics[width=1.0\linewidth]{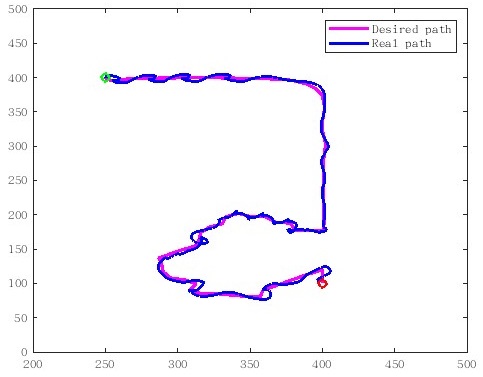}
 \centering
 \caption{A comparison of the desired path and real path in the path-following mission. The start point is represented in green, the goal point is represented in red, the desired path is represented in purple and the real path is represented in blue.}
 \label{desired_real_paths} 
 \end{figure}

\subsection{Path-following mission}
There are several maze-like scenarios in the inshore maritime environment such as the complex, maze-like ports. In this mission, a virtual maze is assumed to exist on the ocean surface in the ROS simulation environment.  In this virtual maze, the selected WAM-V 20 USV platform will follow a path generated by the proposed RRT-GPMP2 motion planner based on the aforementioned fully-autonomous USV navigation framework. 

Fig. \ref{ros_layout_and_demonstration} (a) demonstrates the locations of the virtual maze, start point and goal point, while Fig. \ref{ros_layout_and_demonstration} (b) provides an overview of the ROS simulation environment of the path-following mission. As demonstrated in Fig. \ref{ros_layout_and_demonstration} (a), the positions of the start point and goal point are [400, 100] and [400, 400], respectively. As demonstrated in Fig. \ref{ros_layout_and_demonstration} (b), the WAM-V 20 USV platform is initialised at a random point on the ocean surface and no real obstacle is placed on the ocean surface.

To provide a more intuitive understanding of the path-following mission, the storyboards of the mission from a third-person perspective are shown in Fig. \ref{ros_path_following}, while a comparison of the desired path and real path in the mission is provided in Fig. \ref{desired_real_paths}. As demonstrated in Fig. \ref{desired_real_paths}, although there were some drifts during the path-following process, the USV successfully followed the path generated by the proposed RRT-GPMP2 motion planner.

\section{Conclusion and future research}
\label{Conclusion}
This article introduced an effective motion planner, RRT-GPMP2, to tackle the motion planning problem for mobile robots in complex maze environments. To be more specific, the proposed RRT-GPMP2 motion planner successfully integrates the global planning of the GPMP2 algorithm and the local re-planning of the RRT algorithm. 

Additionally, the proposed RRT-GPMP2 motion planner has been tested in two self-constructed complex maze environments and the simulation results can be summarised as follows: 
\begin{itemize}
    \item The proposed RRT-GPMP2 motion planner can generate a feasible path with higher smoothness compared with the RRT global planning.
    \item The path length of the proposed RRT-GPMP2 motion planner is similar compared with the path length of the RRT global planning.
    \item The total time cost of the RRT local re-planning and GPMP2 global planning in the proposed RRT-GPMP2 motion planner is lower compare with the time cost of the RRT global planning.
\end{itemize}

Lastly, the practicability of the proposed RRT-GPMP2 motion planner has been validated in a high-fidelity maritime simulation environment in ROS. The path generated by the proposed RRT-GPMP2 motion planner is capable of guiding a WAM-V 20 USV to move from a start point to a goal point in a 2-D virtual maritime maze environment.

\subsection{Future research directions}
In terms of future research, some potential directions are proposed as follows:
\begin{itemize}
    \item Expanding the potential application scenarios of the proposed RRT-GPMP2 motion planner by implementing it into 3-D simulation environments.
    \item Testing the proposed RRT-GPMP2 motion planner on other mobile robot platforms such as aerial robots and ground robots.
\end{itemize}

\bibliographystyle{elsarticle-num-names} 
\bibliography{Paper}

\end{document}